\documentclass[12pt]{article} % Anonymized submission
\usepackage{times}
\usepackage{amsmath}
\usepackage{amssymb}
\usepackage{fullpage}
\usepackage{graphicx}
\usepackage{multicol}
\usepackage{color}
\usepackage{indentfirst}
\usepackage{dsfont}
\usepackage[ruled]{algorithm2e}
\usepackage{setspace}
\usepackage{appendix}
\usepackage{comment}
\usepackage[round]{natbib}
\newcommand{\bs}{\boldsymbol}
\allowdisplaybreaks

\title{Combinatorial Multi-Armed Bandits with Filtered Feedback}
\author{James A Grant \\
	\textit{Lancaster University}
	\and
	David S Leslie \\
	\textit{Lancaster University}
	\and
	Kevin Glazebrook \\
	\textit{Lancaster University}
	\and
	Roberto Szechtman\\
	\textit{Naval Postgraduate School}
}

\begin{document}

\maketitle

\begin{abstract}
	Motivated by problems in search and detection we present a solution to a Combinatorial Multi-Armed Bandit (CMAB) problem with both heavy-tailed reward distributions and a new class of feedback, filtered semibandit feedback. In a CMAB problem an agent pulls a combination of arms from a set $\{1,...,k\}$ in each round, generating random outcomes from probability distributions associated with these arms and receiving an overall reward. Under semibandit feedback it is assumed that the random outcomes generated are all observed. Filtered semibandit feedback allows the outcomes that are observed to be sampled from a second distribution conditioned on the initial random outcomes. This feedback mechanism is valuable as it allows CMAB methods to be applied to sequential search and detection problems where combinatorial actions are made, but the true rewards (number of objects of interest appearing in the round) are not observed, rather a filtered reward (the number of objects the searcher successfully finds, which must by definition be less than the number that appear). We present an upper confidence bound type algorithm, Robust-F-CUCB, and associated regret bound of order $\mathcal{O}(\ln(n))$ to balance exploration and exploitation in the face of both filtering of reward and heavy tailed reward distributions.
\end{abstract}

\section{Introduction}
In this paper we present a solution to Combinatorial Multi-Armed Bandit (CMAB) problem with both heavy-tailed reward distributions and filtered semi-bandit feedback. This work is motivated by an application in search and detection (afforded a more detailed description in Section \ref{Motivation}), where an agent sequentially selects combinations of cells to search, aiming to detect some objects in each cell. The number of objects in a given cell in a given round is randomly drawn from a Poisson distribution. The generalisation of previous work on CMAB problems to allow heavy tailed reward distributions is required to accommodate Poisson distributed counts of objects of interest. However due to the imperfect nature of search, this Poisson observation will not necessarily be observed as some events may go undetected. Moreover, the larger the area chosen for search, the less efficient the search will be.  The filtered semibandit feedback allows for the observed outcomes to be a second random outcome, drawn from a distribution conditioned on the \emph{true} or initial outcome. This new feedback class allows us to model the imperfect detection of objects of interest. 

In a CMAB problem with semibandit feedback an agent is faced with $k$ bandit arms representing basic actions and may select some subset of these arms to play at each time step. Each arm $i \in \{1,...,k\}$ has an associated probability distribution $\nu_i$ with finite mean $\mu_i$, both unknown to the agent. Playing a combination of arms $S \subseteq \{1,..,k\}$ reveals random outcomes sampled independently from distributions $\nu_i : i \in S$ and grants the agent a \emph{reward} $R(S)$ which is a function of the random outcomes observed. The agent's goal is to maximise her cumulative reward (or equivalently minimise her cumulative regret) through time.

The work in this paper extends the CMAB framework of \cite{Chen2013} in two directions. Firstly, to one where the underlying probability distributions $\nu_i$ are only restricted to have a bounded moment of order $1+\epsilon$, for $\epsilon \in (0,1]$. Secondly, we introduce \emph{filtered semibandit feedback} where observed outcomes associated with an arm $i$ need not be drawn from $\nu_i$ but can be  \emph{filtered observations} drawn from a related \emph{filtering distribution} $\tilde{\nu}_i$ whose parameters depend on the true outcome $X_{i,t}$ drawn from $\nu_i$ and the combination of arms $S_t$ selected in a given round. The reward received is a function of the filtered observations. As in other versions of the CMAB problem the agent's goal is to maximise her cumulative reward through time.

In addition to introducing this expanded view of the existing CMAB framework, we propose a general class of \emph{upper confidence bound algorithms} for CMAB problems with filtered semibandit feedback, which achieve $\mathcal{O}(\ln n)$ regret subject to the identification of a suitable mean estimator for the specific problem. We include illustrative examples of specific algorithms within this general class for particular distributional families and filtering mechanisms. 

So far as we are aware, the notion of observing filtered rewards in a CMAB problem is a new one and no previous work on algorithms for use under filtered semibandit feedback exists. For CMAB problems with semibandit feedback, the majority of work has focussed on a (relatively) simple CMAB framework where (at most) $m < k$ arms are selected in each round and the overall reward observed is simply the sum of the outcomes generated from these $m$ arms. This configuration is sometimes called Learning with Linear rewards or a Multiple play Bandit and has been considerd by authors including \cite{Gai2012}, \cite{Combes2015}, and \cite{Luedtke2016}. \cite{Chen2013} was the first paper to consider a more general class of reward functions, allowing all functions that satisfy certain smoothness assumptions. It is upon this work that our research is based. \cite{Chen2016a} permits an even less restrictive class of reward functions while \cite{Chen2016b} considers a variant of the usual semibandit feedback where arms not selected in a particular round may still be triggered with a certain probability. We have not attempted to incorporate the two latter innovations in to our work, principally because it was not relevant to our motivating application.

Most Multi-Armed Bandit (MAB) research deals with compact or sub-Gaussian reward distributions, however there are several notable exceptions. In particular \cite{Bubeck2013} present Robust-UCB algorithms suitable for heavy tailed (non sub-Gaussian) reward distributions with a bounded $1+\epsilon$ moment. Extending these algorithms to Robust-CUCB algorithms will be one of our contributions in this work. The Bayes-UCB method of \cite{Kaufmann2012} and KL-UCB method of \cite{Cappe2013} have recently been improved in \cite{Kaufmann2016} to versions with provable regret bounds of optimal order in MAB problems with exponential family rewards. However, as these algorithms are based upon quantile-type UCB indices, rather than UCB indices which take the form of a mean estimate plus an inflation term, the existing analysis from \cite{Chen2013} cannot be so easily exploited and we do not consider a combinatorial extension of these methods in this work.

The rest of the paper is organised as follows. In Section \ref{Motivation} we outline the aforementioned motivating application and justify its link to the CMAB problem. Section \ref{Framework} defines more rigorously our generalisation of the CMAB to include filtering of reward and heavy-tailed reward distributions. Section \ref{Algorithms Section} introduces our main Robust-F-CUCB algorithm for this generalised CMAB problem along with a performance guarantee in the form of a bound on expected regret. We conclude by revisiting the motivating example in light of our theoretical work and providing a short discussion.

\section{Motivating Example - Learning in Search} \label{Motivation}

Our inspiration to study these CMAB problems with filtered semibandit feedback comes from a real world problem in search and detection. In this section we describe this motivating problem and its link to Combinatorial bandits. 

\subsection{Problem Specification}

This research is motivated by the problem of searching for objects over a large area \cite{Stone1976}. The main assumption is that the target objects appear in the search area according to a nonhomogeneous spatial Poisson process. 

Repeated searches are conducted over this search area which is split into a finite number of cells. At each time $t=1,2,...,n$ objects appear according to the Poisson process and the agent selects a subset of the cells to search over (with objects disappearing whether detected or not at time $t+1$). However, the more cells the agent opts to search, the less effective her search can be in any one cell. The key operational question is: How should the cells be searched in order to maximize the expected number of detections over a finite time horizon? We assume that to aid in answering this question, the probability of detecting an object that has appeared in a particular cell given the set of cells the agent opts to search is known. 

If the intensity function of the Poisson process were known, the analyst could formulate a mixed integer linear optimization problem to find the optimal subset of cells to patrol. The challenge for the agent is to come up with a patrolling scheme that judiciously balances exploration and exploitation. Specifically, in this example, a patrolling scheme should take the form of a choice of cells to search in rounds $t=1,2,...,n$, where choices may be made after observing the detections from  previous rounds.

\subsection{Link to Combinatorial Multi-Armed Bandits}

Clearly this problem in search with an unknown intensity function is a sequential decision problem, where the action space in each round is formed of different combinations of cells that the agent may patrol. A CMAB problem is therefore an appropriate model. In each round the agent will choose a set of cells in which to search, so that cells are viewed as bandit arms. Due to the spatio-temporal Poisson process model we specify, the number of objects appearing in a given cell $i$ over a fixed time window will be Poisson distributed with parameter $\mu_i$, independently of the number of objects in other cells. 

However, an added complication comes from the fact that not all objects which appear are detected. Under the choice of combination of arms $S_t$ at time $t$, each object is detected with a certain probability $\gamma_{i,S_t}$ - assumed constant within a round - such that the number of objects observed given the number of objects appearing is Binomially distributed. i.e. if $X_{i,t} \sim Pois(\mu_i)$ is the number of objects appearing in cell $i$ during round $t$, then conditional on $X_{i,t}$ and $S_t$, the number of objects detected $Y_{i,t}$ will have a Binomial distribution such that $Y_{i,t}|X_{i,t},S_t \sim Bin(X_{i,t},\gamma_{i,S_t})$. A consequence of $X_{i,t}$ being Poisson is that the marginal distribution of $Y_{i,t}|S_t$ will follow a $Pois(\gamma_{i,S_t}\mu_i)$ distribution. 

So while there is a clear link between the search problem and the CMAB problem in terms of sequential decision making with a combinatorially structured action space, the original CMAB model of \cite{Chen2013} does not apply directly to the search problem. In the search problem, draws from the underlying reward distribution are not observed. Further, due to the varying detection probabilities from round to round (as different combinations of arms are played) the distribution from which rewards are observed does not remain constant either. Additionally, Poisson rewards have heavier tails than can be accommodated within the framework considered by \cite{Chen2013}. This motivates us to develop an extended CMAB framework allowing for a broader range of underlying reward distributions and a feedback mechanism where the observed rewards are a filtered version of the true outcomes from the underlying distributions. With such a model design, algorithms to approach the search problem can be developed.

\section{Framework} \label{Framework}

In a CMAB problem, an agent is faced with $k$ arms each associated with some unknown, underlying probability distribution $\nu_i$ with expectation $\mu_i$. At each time step $t=1,2,...$ the agent selects a combination of arms $S_t$ from a set of possible combinations $\mathcal{S} \subseteq \mathcal{P}\big(\{1,...,k\}\big)$ where $\mathcal{P}\big(\{1,...,k\}\big)$ denotes the power set of the set of arms. When a combination of arms is selected in a round, we say that all the arms within that combination have been played in the round. Letting $T_{i,t}=\sum_{j=1}^t\mathds{I}\{i \in S_j\}$ denote the number of times an arm is played in the first $t$ rounds, we introduce the \emph{filtered semi bandit feedback} framework as follows.

When a combination of arms $S_t$ is selected in round $t$, a random outcome $X_{i,T_{i,t}}$ is generated (independently) from underlying distribution $\nu_i$ for each $i \in S_t$. However, these outcomes remain unobserved. Instead, for each $i \in S_t$, a \emph{filtered observation} $Y_{i,T_{i,t}}$ is drawn from a \emph{filtering distribution} $\tilde{\nu}_{i,T_{i,t}}=\tilde{\nu}_i(X_{i,T_{i,t}},S_t)$ conditioned on the random outcome from the underlying distribution and the combination of arms played. These filtered observations \emph{are} seen by the agent. Let $\mathbf{X}_{S_t}$ and $\mathbf{Y}_{S_t}$ respectively denote the vectors of true outcomes and filtered observations in round $t$ where combination of arms $S_t$ is selected. 

In addition to observing $\mathbf{Y}_{S_t}$, playing the combination of arms $S_t$ grants the agent a reward $R(\mathbf{Y}_{S_t})$ which is a function of the filtered observations (and thus is a random variable). The expectation of the reward obtained by playing combination of arms $S_t$, with respect to a particular vector of underlying means $\bs\mu$, is denoted $r_{\bs\mu}(S_t)=\mathds{E}(R(\mathbf{Y}_{S_t})|S_t)$. The function $r_{\bs\mu}: \mathcal{S} \rightarrow \mathds{R}$ is referred to as a \emph{reward function}.

One example of a filtering model is the binomial filtering of discrete non-negative integer data, as seen in the search example of Section \ref{Motivation}. In such a model, $Y_{i,T_{i,t}}|X_{i,T_{i,t}},S_t$ follows a $Bin(X_{i,T_{i,t}},\gamma_{i,S_t})$ distribution where $\gamma_{i,S_t}$ is a success probability dependent on the combination of arms played.

Filtered semibandit feedback can be contrasted with bandit, semibandit and full information feedbacks where there is no filtering, or in our terms where the filtering distributions are such that $\mathbf{Y}_{S_t}=\mathbf{X}_{S_t}$ with probability 1 (so true outcomes and filtered observations can be treated as the same thing). In bandit feedback, only $R(\mathbf{X}_{S_t})$ is observed. In semibandit feedback $R(\mathbf{X}_{S_t})$ and $\mathbf{X}_{S_t}$ are observed. In full information feedback $R(\mathbf{X}_{S_t})$, $\mathbf{X}_{S_t}$, and a draw from $\nu_i$ for $i \notin S_t$ are observed. Our model applies filtering to the semibandit feedback case. We do not consider filtered variants of bandit or full information feedback. We note that the classical stochastic Multi-Armed Bandit (MAB) problem is a special case of the CMAB problem with (non-filtered) bandit or semibandit feedback, where $\mathcal{S}=\big\{\{1\},...,\{k\}\big\}$ and the reward observed is simply equal to the observation $X_{i,T_{i,t}}$ drawn from the arm $i$ selected.

A CMAB problem with filtered semibandit feedback is therefore defined by a set of underlying probability distributions $\bs\nu=(\nu_1,...,\nu_k)$ with means $\bs\mu=(\mu_1,...,\mu_k)$, a set of possible combinations $\mathcal{S}$, a reward function $r_{\bs\mu}(\cdot)$, and a set of filtering distributions $\tilde{\bs\nu}=(\tilde{\nu}_1,...,\tilde{\nu}_k)$ with variable parameters. To aid in the analysis in this paper, we make assumptions on the expected reward $r_{\bs\mu}(S)$ as in \cite{Chen2013}.  \\

\noindent \emph{\textbf{Assumption 1} - Monotonicity}: 

\noindent The expected reward of playing any combination of arms $S \in \mathcal{S}$ is monotonically nondecreasing with respect to the expectation vector, i.e. if for all $i \in \{1,...,k\},$ $\mu_i \leq \mu_i'$, we have $r_{\bs\mu}(S) \leq r_{\bs\mu'}(S)$ for all $S \in \mathcal{S}$. \\

\noindent \emph{\textbf{Assumption 2} - Bounded Smoothness}: 

\noindent There exists a strictly increasing function $f(\cdot)$ called a \emph{bounded smoothness function}, such that for any two expectation vectors $\bs\mu$ and $\bs\mu'$ with $\max_{i \in S}|\mu_i - \mu_i'|\leq \Lambda$ we have $|r_{\bs\mu}(S)-r_{\bs\mu'}(S)|\leq f(\Lambda)$. \\

With these assumptions in place we will be able to construct bounds on the performance of UCB-type algorithms for CMAB problems with filtered semibandit feedback. A CMAB algorithm will, in a round $t$, consider the rewards observed in previous rounds and select a combination of arms $S_t$ to be played. Its objective is to maximise cumulative expected reward over $n$ rounds, $\mathds{E}\big(\sum_{t=1}^n r_{\bs\mu}(S_t)\big)$ - where the expectation is taken with respect to the actions selected by the algorithm.

We investigate the performance of UCB type algorithms for the CMAB problems. Typically, UCB algorithms make decisions based on indices formed by adding an inflation term to a data-driven estimator of the underlying mean $\mu_i$. Successful algorithms are obtained by selecting the inflation term appropriately to match the convergence rate of the mean estimator thereby encouraging an appropriate balance of exploration and exploitation. 

In simple CMAB or MAB problems with bounded or sub-Gaussian reward distributions, an empirical mean has convergence of a suitable rate to yield UCB algorithms with $\mathcal{O}(\ln n)$ bounded regret. However with non-sub-Gaussian (or \emph{heavy tailed}) reward distributions the empirical mean lacks this same rate of convergence. As in \cite{Bubeck2013}, we turn to more robust mean estimators to find the correct convergence rate. A further challenge is that our mean estimators must be based on observations from filtered distributions but converge to the mean of the underlying distributions. We seek estimators $\hat{\mu}(Y_{i,1},...,Y_{i,n})$ of $\mu_i$ based on filtered observations $Y_{i,1},...,Y_{i,n}$ which satisfy the following assumption for the relevant distributions in the particular CMAB problems we consider. \\

\noindent \emph{\textbf{Assumption 3} - Concentration of Mean Estimator}: 

\noindent
The mean estimator $\hat{\mu}_{i,n}=\hat{\mu}(Y_{i,1},...,Y_{i,n})$ is such that for positive parameter $\epsilon \in (0,1]$, positive values $c,v$, and independent random variables $Y_{i,1},...,Y_{i,n}$ drawn from filtering distributions $\tilde{\nu}_{i,1},...,\tilde{\nu}_{i,n}$ we have for all $n\geq 1$ and  $\delta \in (0,1)$ \begin{align}
\mathds{P}\bigg(\hat{\mu}_{i,n} \geq \mu_i + v^{\frac{1}{1+\epsilon}}\Big(\frac{c\ln(1/\delta)}{n}\Big)^{\frac{\epsilon}{1+\epsilon}}\bigg) & \leq \delta \label{Assumption 3 Pt. 1} \\
\mathds{P}\bigg({\mu}_i \geq \hat{\mu}_{i,n} + v^{\frac{1}{1+\epsilon}}\Big(\frac{c\ln(1/\delta)}{n}\Big)^{\frac{\epsilon}{1+\epsilon}}\bigg) & \leq \delta. \label{Assumption 3 Pt. 2}
\end{align}

\section{Robust-F-CUCB} \label{Algorithms Section}

For the stochastic CMAB with filtered semibandit feedback as introduced in Section \ref{Framework}, we propose the Robust-F-CUCB algorithm, described in Algorithm 1. The Robust-F-CUCB algorithm is both a generalisation and combination of the Robust-UCB algorithm of \cite{Bubeck2013} and CUCB algorithm of \cite{Chen2013}. This section proceeds as follows: We begin by introducing the necessary language and notation to express performance guarantees in the form of regret bounds. We then present our Robust-F-CUCB algorithm for a general mean estimator satisfying Assumption 3, before considering versions of the algorithm tailored to more specific reward and filtering distributions. With each version of the algorithm we also present a regret bound.

\subsection{Regret Notation}

The regret of a CMAB algorithm in $n$ rounds with respect to an expectation vector $\bs\mu$ can be written \begin{equation}
Reg_{n,\bs\mu} = n\cdot\text{opt}_{\bs\mu} - \mathds{E}\bigg(\sum_{t=1}^n r_{\bs\mu}(S_t)\bigg). \label{Regret Definition}
\end{equation}
where $\text{opt}_{\bs\mu}=\max_{S \in \mathcal{S}}r_{\bs\mu}(S)$ denotes the highest attainable expected reward in a single round of the CMAB problem with respect to a expectation vector $\bs\mu$, and the expectation in (\ref{Regret Definition}) is taken with respect to the actions selected by the algorithm. The aim to maximise expected reward is equivalent to minimising regret. The quality of an algorithm is usually measured by determining an analytical bound on $Reg_{n,\bs\mu}$, the order of which is the principal consideration. Algorithms with bounds of $\mathcal{O}(\ln n)$ are said to be of optimal order in CMAB and MAB problems. 

In line with Chen et al. we define \begin{align*}
\Delta_{max}&=\text{opt}_{\bs\mu}-\min_{S \in \mathcal{S}}\{r_{\bs\mu}(S) \enspace | \enspace r_{\bs\mu}(S) \neq \text{opt}_{\bs\mu}\} \\
\Delta_{min}&=\text{opt}_{\bs\mu}-\max_{S \in \mathcal{S}}\{r_{\bs\mu}(S) \enspace | \enspace r_{\bs\mu}(S) \neq \text{opt}_{\bs\mu}\} 
\end{align*} The quantity $\Delta_{max}$ is then the difference in expected reward between an optimal combination of arms and the worst possible combination of arms, while $\Delta_{min}$ is the difference in expected reward between an optimal combination of arms and the closest to optimal suboptimal combination of arms. These quantities will be important in defining bounds on expected regret.

\subsection{General Algorithm Statement and Regret Bound}
We first describe the Robust-F-CUCB algorithm for a general mean estimator satisfying Assumption 3, before considering more specific results later in this section. Like the CUCB algorithm, our Robust-F-CUCB algorithm consists firstly of an initialisation stage where a combination of arms containing each arm is played, to initialise mean estimates and $T_{i,t}$ counters. Thereafter, in each round, upper confidence bounds (UCBs) are calculated for each arm and these UCBs are passed to a combinatorial optimisation to identify the best combination of arms to play from our optimistic perspective. The approach is presented in full detail in Algorithm 1.

\begin{algorithm}[]
	\label{RobustFCUCBAlgorithm}
	\hrule
	\vspace{0.1cm}
	\caption{Robust-F-CUCB}
	\vspace{0.1cm}
	\hrule
	\vspace{0.1cm}
	\textbf{Inputs:} Parameter $\epsilon \in (0,1]$, positive constants $c,v$, and mean estimator $\hat{\mu}_{i,n}$ satisfying Assumption 3. \\
	
	\textbf{Initialisation Phase:} \\
	For each arm $i$, play an arbitrary combination of arms $S \in \mathcal{S}$ such that $i \in S$. \\ 
	Set $t$ $\leftarrow$ $k$. \\
	
	\textbf{Loop Phase:} \\
	For $t = k+1, k+2,....$ \\
	\hphantom{a} $\quad$ For each arm $i$, calculate $\bar{\mu}_{i,T_{i,t-1},t} = \hat{\mu}_{i,T_{i,t-1}} + v^\frac{1}{1+\epsilon}\bigg(\frac{c\ln t^3}{T_{i,t-1}}\bigg)^{\frac{\epsilon}{1+\epsilon}}$. \\
	\hphantom{a} $\quad$ Play a combination of arms $S_t$ with $r_{\bar{\bs\mu}_t}(S_t)=\max_{S\in \mathcal{S}}r_{\bar{\bs\mu}_t}(S)=\text{opt}_{\bar{\bs\mu}_t}$. \\
	\hphantom{a} $\quad$ Update $T_i$ and $\hat\mu_i$ for all $i \in S_t$. 

	\hrule
\end{algorithm}

\noindent
The following theorem provides our performance bound for the Robust-F-CUCB algorithm. \\

\noindent
\emph{\textbf{Theorem 1:}} 
Let $\epsilon \in (0,1]$ and let $\hat{\mu}_{i,n}$ be a mean estimator. Suppose that the underlying distributions $\nu_1,...,\nu_k$ and filtering distributions $\tilde{\nu}_1,...,\tilde{\nu}_k$ are such that the mean estimator satisfies Assumption 3 for all $i=1,...,k$. Then the regret of the Robust-F-CUCB policy satisfies \begin{equation}
Reg_{n,\bs\mu} \leq \Bigg(3cv^{\frac{1}{\epsilon}}\bigg(\frac{2}{f^{-1}(\Delta_{min})}\bigg)^{\frac{1+\epsilon}{\epsilon}}\ln n + \frac{\pi^2}{3} +1 \Bigg)\cdot k \cdot \Delta_{max}. \label{Theorem 1 Bound}
\end{equation}

\noindent
\emph{Proof:} Since the mean estimators are assumed to satisfy Assumption 3, the proof is an adaptation of those given by \cite{Chen2013} and \cite{Bubeck2013} and is given in Appendix \ref{Theorem1Proof}. \\

A particular instance of the Robust-F-CUCB algorithm can be defined by a specific mean estimator and particular values of $\epsilon,$ $c,$ $v$. In the remainder of this section, we consider particular cases of the CMAB problem and particular Robust-F-CUCB algorithms which are appropriate to these problems.

\subsection{Semibandit Feedback with Heavy Tails}

Firstly, we consider a CMAB problem without filtering - i.e. where the filtering distributions are such that $Y_{i,t}=X_{i,t}$ for all $i$ and $t$ and we simply have semibandit feedback. In this situation we are still considering a model not previously studied in the literature as we have permitted a more general class of reward distribution whose support is not solely contained within $[0,1]$. For this problem class we propose the Robust-F-CUCB algorithm with truncated empirical mean - a direct extension of the Robust-UCB policy with truncated empirical mean specified by \cite{Bubeck2013}. The truncated empirical mean, given some parameters $u>0$, and $\epsilon \in (0,1]$, and based on observations $X_1,...,X_n$ is defined as \begin{equation}
\hat{\mu}_{i,n}^{Trunc}=\frac{1}{n}\sum_{t=1}^n X_t \mathds{I}\bigg\{|X_t|\leq \bigg(\frac{ut}{\ln(t)}\bigg)^{\frac{1}{1+\epsilon}}\bigg\}. \label{TruncatedEmpiricalMean}
\end{equation}

The following Proposition provides a bound on the regret of the Robust-F-CUCB algorithm where the underlying distributions $\bs\nu$ have suitably bounded $1+\epsilon$ moments. The Robust-F-CUCB algorithm with truncated empirical mean works for this problem class because in \cite{Bubeck2013} the truncated empirical mean has already been shown to satisfy Assumption 3. \\

\noindent
\emph{\textbf{Proposition 2}}: 
Let $\epsilon \in (0,1]$ and $u > 0$. Let the reward distributions $\nu_1,...,\nu_k$ satisfy \begin{displaymath}
\mathds{E}_{X \sim \nu_i}|X_i|^{1+\epsilon} \leq u \quad \forall \enspace i \in \{1,...,k\}.
\end{displaymath}
Then the regret of the Robust-F-CUCB algorithm used with the truncated empirical mean estimator defined in (\ref{TruncatedEmpiricalMean}) satisfies \begin{equation}
Reg_{n,\bs\mu} \leq \Bigg(12(4u)^{\frac{1}{\epsilon}}\bigg(\frac{2}{f^{-1}(\Delta_{min})}\bigg)^{\frac{1+\epsilon}{\epsilon}}\ln n + \frac{\pi^2}{3} + 1\Bigg)\cdot k \cdot \Delta_{max}. \label{Lemma2RegretBound}
\end{equation}

\noindent
\emph{Proof}: \cite{Bubeck2013} shows that Assumption 3 holds with $c=4$ and $v=4u$. The main result then follows from Theorem 1.

\subsection{Binomially filtered Poisson rewards} \label{BinFilt}

We now wish to consider a CMAB problem with filtering. As we mentioned in Section 3, one possible filtering framework is the Binomial filtering of count data, i.e. where if in round $t$ $X_{i,T_{i,t}}$ is a draw from $\nu_i$ and $S_t$ is the combination of arms selected then the filtering distribution for arm $i$, $\tilde{\nu}_i(X_{i,T_{i,t}},S_t)$ is $Bin(X_{i,T_{i,t}},\gamma_{i,S_t})$. We also mentioned that if $\nu_i$ follows a Poisson distribution with parameter $\mu_i$ then the marginal distribution of the filtered observation $Y_{i,T_{i,t}}|S_t$ will be Poisson with parameter $\gamma_{i,S_t}\mu_i$. We consider this example, with the additional assumption that for some $\gamma_{min}>0$ we have $\gamma_{i,S}>\gamma_{min}$ for all $i$ and $S \in \mathcal{S}$.

To define a Robust-F-CUCB algorithm that satisfies the logarithmic order regret bound for this problem, we must have a mean estimator which satisfies Assumption 3 for the filtering distributions specified above. Consider the following \emph{filtered} truncated empirical mean estimator, and the associated Lemma, demonstrating that a version of Assumption 3 holds for this estimator when applied to Poisson reward distributions. \\

\noindent
\emph{\textbf{Lemma 3}}: Let $\delta\in (0,1)$ and $\mu_{max}>0$, and define $u_{max}=\mu_{max}^2 + \mu_{max}$. Consider a series of filtered Poisson observations $Y_{i,1},...,Y_{i,n}$ with means $\gamma_{i,1}\mu_i,...,\gamma_{i,n}\mu_i$ where $\gamma_{i,t} \in (\gamma_{min},1]$ for $t=1,...,n$ and $\gamma_{min} > 0$. Consider the filtered truncated empirical mean estimator, \begin{equation}
\hat{\mu}_{i,n}^{TruncF} = \frac{1}{n}\sum_{t=1}^n \frac{Y_{i,t}}{\gamma_{i,t}} \mathds{I}{\Bigg\{Y_{i,t} \leq \gamma_{i,t} \sqrt{\frac{u_{max}t}{\ln \delta^{-1}}}\Bigg\}}. \label{FiltTruncMean}
\end{equation} If $\mu_i \leq \mu_{max}$ then \begin{align}
\mathds{P}\Bigg(\hat{\mu}_{i,n}^{TruncF} \geq \mu_i + \Big(\frac{2}{\gamma_{min}} + \sqrt{\frac{2}{\gamma_{min}}} +\frac{1}{3}\Big)\sqrt{\frac{u_{max} \ln \delta^{-1}}{n}}\Bigg) &\leq \delta, \label{Lemma4Pt1} \\ 
\mathds{P}\Bigg(\mu_i \geq \hat{\mu}_{i,n}^{TruncF} + \Big(\frac{2}{\gamma_{min}} + \sqrt{\frac{2}{\gamma_{min}}} +\frac{1}{3}\Big)\sqrt{\frac{u_{max} \ln \delta^{-1}}{n}}\Bigg) &\leq \delta. \label{Lemma4Pt2}
\end{align} \\

\noindent
We present a proof of Lemma 3 in Appendix \ref{Lemma3Proof}. Proposition 4 below specifies the regret bound which holds if the filtered truncated empirical mean estimator is used to define a Robust-F-CUCB algorithm and this algorithm is applied to the CMAB problem with the filtering structure defined above.  \\

\noindent
\emph{\textbf{Proposition 4}}: Let $\epsilon=1$ and $\mu_{max}>0$. Let the reward distributions $\nu_1,...,\nu_k$ be Poisson satisfying $\mu_i \leq \mu_{max}$ for $i=1,...,k$. Let the filtering distributions $\tilde{\nu}_1,...,\tilde{\nu}_k$ be Binomial as described above. Then the regret of the Robust-F-CUCB algorithm used with the filtered truncated empirical mean estimator defined in (\ref{FiltTruncMean}) satisfies \begin{equation}
Reg_{n,\bs\mu} \leq \Bigg(\frac{12(\mu_{max}^2+\mu_{max})\Big(\frac{2}{\gamma_{min}} + \sqrt{\frac{2}{\gamma_{min}}} +\frac{1}{3}\Big)^2}{\Big(f^{-1}(\Delta_{min})\Big)^2}\ln n + \frac{\pi^2}{3} + 1\Bigg)\cdot k \cdot \Delta_{max}. \label{Lemma4RegretBound}
\end{equation}

\noindent
\emph{Proof}: Lemma 3 shows that Assumption 3 holds with $\epsilon=1$, $c=u_{max}$ and $v=\big(\frac{2}{\gamma_{min}} + \sqrt{\frac{2}{\gamma_{min}}} +\frac{1}{3}\big)^2$. The main result then follows from Theorem 1.
 
\section{Discussion}
Within this paper we have presented a generalisation (in two senses) of the Combinatorial Multi-Armed Bandit framework, by considering unbounded reward distributions and filtered semibandit feedback. Our Robust-F-CUCB algorithm, presented in a general form, can be shown to have an associated logarithmic order bound on regret and we have specified this bound for particular CMAB problem instances. In particular we have shown that in a filtering free context, the truncated mean estimator  can be used to provide an algorithm for a CMAB problem with heavy tails with a logarithmic order bound on regret. We developed a generalisation of the truncated mean estimator to deal with binomially filtered Poisson data and showed that for this class of data, it has the required concentration properties - a result which could of course be applied in the study of other problems, not just bandits.

We can apply the Robust-F-CUCB algorithm with filtered truncated empirical mean discussed in Section \ref{BinFilt} in the sequential search problem as long as we have knowledge of some upper bound $\mu_{max}$ such that the average rate of the underlying Poisson process in each cell is below the upper bound in each round. As the reward function in this problem is the expected number of detected events, $r_{\bs\mu}(S_t)=\sum_{i \in S_t} \gamma_{i,S_t}\mu_i$, and $\gamma_{min} \leq \gamma_{i,S_t} \leq 1$ for all $i$ and $S_t \in \mathcal{S}$, Assumption 1 (monotonicity) holds and Assumption 2 (bounded smoothness) holds for a bounded smoothness function $f(\Lambda)=k\Lambda$. Thus the we can bound the regret of the Robust-F-CUCB as $\mathcal{O}(k^3\ln n)$. 

We note also, that we could readily extend this work to a more complex application where there are multiple agents searching collaboratively. If multiple searchers were each to select a combination of cells to search (in such a way that the combinations do not overlap), one could still identify a best combination in the full information problem by formulating and solving a Integer Linear Program. In the sequential decision variant of the problem would still observe filtered rewards from the multiple combinations of cells in each round and could still apply an Upper Confidence Bound algorithm to balance exploration and exploitation.

The key difference with this more complex application is that the mapping between combinations of arms and an interpretable allocations of searchers would not be one-to-one. The added combinatorial aspect of the problem means that, while a combination of arms would be interpretable as the union of the sets of cells picked by the different searchers, different sets of sets could lead to the same overall combination of arms. Therefore, crucially, a combination of arms may have multiple different sets of filtering distributions associated with it, and the most appropriate \emph{way to play} the combination of arms may vary as the arm indices do. So, to approach this more complex version of the problem, a definition of the set of possible combinations $\mathcal{S}$ that includes labellings of the partitions within combinations of cells $S \in \mathcal{S}$ is required.

We note that it would be possible to improve the leading order coefficients of all our regret bounds by applying the more sophisticated analysis used in the proof of Theorem 1 of \cite{Chen2013} to our Robust-F-CUCB framework. Said analysis would improve on our presented analysis by noting the discrepancy between defining sufficiency of sampling with respect to $\Delta_{min}$ but bounding regret with respect to $\Delta_{max}$. The more sophisticated analysis would remain usable in our more complex framework, as any intricacies due to filtering can be truly captured within the concentration inequality based step, which yields the $\frac{\pi^2}{3}k\Delta_{max}$ term, a step which the more sophisticated analysis does not alter. We have refrained from making these improvements in this work, as they do not affect the order of the bound and the omission permits an easier explanation of our key results. Furthermore although we did not directly consider the more general $(\alpha, \beta)$-\emph{approximation regret} considered by Chen et al. (which allows for the CMAB algorithm to be a randomised algorithm with a small failure probability), the results presented in our paper can be trivially generalised to incorporate this by reintroducing the $\alpha$ and $\beta$ parameters which we have effectively fixed to equal 1. 

\section*{Acknowledgements}{We gratefully acknowledge the support of the EPSRC funded EP/L015692/1 STOR-i Centre for Doctoral Training}

\bibliographystyle{apalike}
\bibliography{coltpaperrefs}

\appendix

\section{Proof of Theorem 1} \label{Theorem1Proof}
\noindent
\emph{Proof of Theorem 1:}

%\noindent
For each arm maintain $T_{i,t}$ as the a count of the number of times arm $i$ has been played in the first $t$ rounds. We also maintain a second set of counters $\{N_i\}_{i=1}^k$, one associated with each arm. These counters, which collectively count the number of suboptimal plays, are updated as follows. Firstly, after the $k$ initialisation rounds set $N_{i,k}=1$ for all $i \in \{1,...,k\}$. Thereafter, in each round $t >k$, let $S_t$ be the combination of arms played in round $t$ and let $i=\arg\min_{j \in S_t}N_{j,t}$, if $i$ is non-unique then we choose randomly from the minimising set. If $r_{\bs\mu}(S_t) \neq \text{opt}_{\bs\mu}$ then we increment $N_{i}$, i.e. set $N_{i,t}=N_{i,t-1}+1$, . The key results of these updating rules are that $\sum_{i=1}^k N_{i,t}$ provides an upper bound on the number of suboptimal plays in $t$ rounds and $T_{i,t}\geq N_{i,t}$ for all $i$ and $t$.

Define \begin{displaymath}
l_t = 3cv^{\frac{1}{\epsilon}}\bigg(\frac{2}{f^{-1}(\Delta_{min})}\bigg)^{\frac{1+\epsilon}{\epsilon}}\ln(t).
\end{displaymath} 

We consider a round $t$ in which $S_t: r_{\bs\mu}(S_t) \neq \text{opt}_{\bs\mu}$ is selected and counter $N_i$ of some arm $i \in S_t$ is updated. We have \begin{align}
\sum_{i=1}^k N_{i,n} - k &= \sum_{t=k+1}^n \mathds{I}\{r_{\bs\mu}(S_t) \neq \text{opt}_{\bs\mu}\}  \nonumber\\
\Rightarrow \enspace \sum_{i=1}^k N_{i,n} - k\cdot(l_n+1) &= \sum_{t=k+1}^n \mathds{I}\{r_{\bs\mu}(S_t) \neq \text{opt}_{\bs\mu}\}-kl_n \nonumber \\
&\leq \sum_{t=k+1}^n \sum_{i=1}^k \mathds{I}\{r_{\bs\mu}(S_t) \neq \text{opt}_{\bs\mu}, N_{i,t} > N_{i,t-1}, N_{i,t-1} > l_n\} \nonumber \\
&\leq \sum_{t=k+1}^n \sum_{i=1}^k \mathds{I}\{r_{\bs\mu}(S_t) \neq \text{opt}_{\bs\mu}, N_{i,t} > N_{i,t-1}, N_{i,t-1} > l_t\} \nonumber \\
&= \sum_{t=k+1}^n \mathds{I}\{r_{\bs\mu}(S_t) \neq \text{opt}_{\bs\mu}, N_{i,t-1} > l_t \enspace \forall i \in S_t\} \label{Indicator1} \\
&\leq \sum_{t=k+1}^n \mathds{I}\{r_{\bs\mu}(S_t) \neq \text{opt}_{\bs\mu},  T_{i,t-1} > l_t \enspace \forall i \in S_t\} \label{Indicator2}
\end{align}
Here, the initial equations come from the updating rules for counters. The first inequality holds because there are at most $kl_n$ occasions where the specified conditions do not hold - i.e. once each counter has been updated $l_n$ times, none of the counters will be $< l_n$. The second inequality is true because $l_t \leq l_n$ for $t\leq n$, and equation (\ref{Indicator1}) holds because of our rule that we always update only one of the smallest counters in the selected combination of arms. The final inequality follows from $N_{i,t} \leq T_{i,t}$. 

We wish to show $\mathds{P}(r_{\bs\mu}(S_t) \neq \text{opt}_{\bs\mu}, T_{i,t-1}>l_t \enspace \forall i \in S_t) \leq 2kt^{-2}$, so that the summation in (\ref{Indicator2}) converges. As a consequence of Assumption 3, for any arm $i=1,..,k$ we have: \begin{align}
\mathds{P}\Bigg(|\hat{\mu}_{i,T_{i,t-1}}-\mu_i| \geq v^{\frac{1}{1+\epsilon}}\bigg(\frac{c\ln t^3}{T_{i,t-1}}\bigg)^{\frac{\epsilon}{1+\epsilon}}\Bigg) 
&=\sum_{s=1}^{t-1} \mathds{P}\Bigg(\Bigg\{|\hat{\mu}_{i,s}-\mu_i| \geq v^{\frac{1}{1+\epsilon}}\bigg(\frac{c\ln t^3}{s}\bigg)^{\frac{\epsilon}{1+\epsilon}},T_{i,t-1}=s\Bigg\}\Bigg) \nonumber \\
&\leq \sum_{s=1}^{t-1} \mathds{P}\Bigg(|\hat{\mu}_{i,s}-\mu_i| \geq v^{\frac{1}{1+\epsilon}}\bigg(\frac{c\ln t^3}{s}\bigg)^{\frac{\epsilon}{1+\epsilon}}\Bigg) \nonumber \\
&\leq t\cdot 2t^{-3} \leq 2t^{-2}. \label{ProbBound}
\end{align}

Define a random variable $\Lambda_{i,t} = v^{\frac{1}{1+\epsilon}}\bigg(\frac{c\ln t^3}{T_{i,t-1}}\bigg)^{\frac{\epsilon}{1+\epsilon}}$ and event $E_t =\{|\hat\mu_{i,T_{i,t-1}} - \mu_i| \leq \Lambda_{i,t}, \forall \enspace i=1,..,k\}$. It is clear, by a union bound on Eq. (\ref{ProbBound}) that $\mathds{P}(\neg E_t) \leq 2kt^{-2}$. In the loop phase of the Robust-F-CUCB algorithm we have $\bar\mu_{i,t} - \hat\mu_{i,T_{i,t-1}} = \Lambda_{i,t}$. Thus, $E_t$ implies $\bar{\mu}_{i,t}\geq \mu_i$ for all $i$. 

Let $\Lambda=v^{\frac{1}{1+\epsilon}}\bigg(\frac{c\ln t^3}{l_t}\bigg)^{\frac{\epsilon}{1+\epsilon}}$ (not a random variable) and define $\Lambda_t = \max\{\Lambda_{i,t}|i \in S_t\}$ (which is a random variable). The following results can then be written: \begin{align}
E_t \Rightarrow  \enspace |\bar{\mu}_{i,t}-\mu_i| \leq 2&\Lambda_t \enspace \forall i \in S_t\label{EventImply}\\
\{r_{\bs\mu}(S_t) \neq \text{opt}_{\bs\mu} , T_{i,t-1} > l_t \enspace \forall i \in S_t \} \Rightarrow \Lambda > &\Lambda_t \label{EventImply2}
\end{align}
which follow from the definitions of the various $\Lambda$ terms.

We can then present the following derivation, true if $\{E_t,r_{\bs\mu}(S_t) \neq \text{opt}_{\bs\mu}, T_{i,t-1} > l_t \enspace \forall i \in S_t\}$ holds: \begin{align*}
r_{\bs\mu}(S_t) + f(2\Lambda) &> r_{\bs\mu}(S_t)+f(2\Lambda_t),  \\
&\geq r_{\bar{\bs\mu}_t}(S_t) = \text{opt}_{\bar{\bs\mu}_t}, \\
&\geq r_{\bar{\bs\mu}_t}(S^*_{\bs\mu}), \\
&\geq r_{\bs\mu}(S^*_{\bs\mu}) = \text{opt}_{\bs\mu},
\end{align*} where $S^*_{\bs\mu}$ is an combination of arms with optimal expected reward with respect to the true mean vector. The first inequality follows from the monotonicity of the bounded smoothness function specified in Assumption 2 and Eq. (\ref{EventImply2}). The second is a result of the bounded smoothness property of Assumption 2 and Eq. (\ref{EventImply}). The third inequality follows from the definition of $\text{opt}_{\bar{\bs\mu}_t}$ and the fourth from the monotonicity of $r_{\bs\mu}(S)$ assumed in Assumption 1 and the result that $E_t \Rightarrow \bar{\bs\mu}_t \geq \bs\mu$. In summary, this derivation says that if $\{E_t,r_{\bs\mu}(S_t) \neq \text{opt}_{\bs\mu}, \forall i \in S_t, T_{i,t-1} > l_t\}$ holds then \begin{equation}
r_{\bs\mu}(S_t)+f(2\Lambda) > \text{opt}_{\bs\mu}. \label{DerivResult}
\end{equation}
The definitions of $l_t$ and $\Lambda$ mean that $f(2\Lambda)= \Delta_{min}$, and we can rewrite Eq. (\ref{DerivResult}) as \begin{equation}
r_{\bs\mu}(S_t) + \Delta_{min} > \text{opt}_{\bs\mu}.
\end{equation}
This however, is a direct contradiction of the definition of $\Delta_{min}$ and the assumption that $r_{\bs\mu}(S_t) \neq \text{opt}_{\bs\mu}$. This means that $\mathds{P}(\{E_t,r_{\bs\mu}(S_t) \neq \text{opt}_{\bs\mu},  T_{i,t-1} > l_t \enspace \forall i \in S_t\})=0$ and thus \begin{displaymath}
\mathds{P}(\{r_{\bs\mu}(S_t) \neq \text{opt}_{\bs\mu},  T_{i,t-1} > l_t \enspace \forall i \in S_t\}) \leq \mathds{P}(\neg E_t) \leq 2kt^{-2}
\end{displaymath}
as derived previously. 

From (\ref{Indicator2}) we can thus write: \begin{align}
\mathds{E}\bigg(\sum_{i=1}^k N_{i,n}\bigg) &\leq k(l_n+1) + \sum_{t=k+1}^n \mathds{P}\big(r_{\bs\mu}(S_t) \neq \text{opt}_{\bs\mu},  T_{i,t-1} > l_t \enspace \forall i \in S_t\big) \nonumber \\
&\leq k(l_n+1) + \sum_{t=1}^n \frac{2k}{t^{2}} \nonumber \\
&\leq \frac{2^{\frac{1+\epsilon}{\epsilon}} \cdot c \cdot k \cdot v^{\frac{1}{\epsilon}} \cdot \ln n^3}{(f^{-1}(\Delta_{min}))^{\frac{1+\epsilon}{\epsilon}}}+ \bigg(\frac{\pi^2}{3} + 1\bigg) \cdot k. \label{BadRoundBound}
\end{align} 
Since the expected reward from playing a suboptimal combination of arms is at most $\Delta_{max}$ from $\text{opt}_{\bs\mu}$ we can trivially reach the required result by assuming the suboptimal rounds are all as far from optimality as they could be. $\square$

\section{Proof of Lemma 3} \label{Lemma3Proof}

\noindent
\emph{Proof of Lemma 3}: The  proof will show (\ref{Lemma4Pt2}) to be true, and then proving (\ref{Lemma4Pt1}) is just a simple modification of the same steps. Define $B_t = \sqrt{\frac{u_{max} t}{\ln \delta^{-1}}}$. We have \begin{align}
\mu_i - \hat{\mu}_{i,n}^{TruncF} &= \frac{1}{n}\sum_{t=1}^n \bigg(\mu_i - \frac{Y_{i,t}}{\gamma_{i,t}}\mathds{I}{\Big\{Y_{i,t} \leq \gamma_{i,t}B_t\Big\}}\bigg) \nonumber \\
&= \frac{1}{n}\sum_{t=1}^n \Bigg(\mathds{E}\bigg(\frac{Y_{i,t}}{\gamma_{i,t}}\bigg) - \mathds{E}\bigg(\frac{Y_{i,t}}{\gamma_{i,t}}\mathds{I}{\Big\{Y_{i,t} \leq \gamma_{i,t}B_t\Big\}}\bigg)\Bigg) \nonumber \\
&\quad \quad \quad \quad + \frac{1}{n}\sum_{t=1}^n \Bigg(\mathds{E}\bigg(\frac{Y_{i,t}}{\gamma_{i,t}}\mathds{I}{\Big\{Y_{i,t} \leq \gamma_{i,t}B_t\Big\}}\bigg) - \frac{Y_{i,t}}{\gamma_{i,t}}\mathds{I}{\Big\{Y_{i,t} \leq \gamma_{i,t}B_t\Big\}}\Bigg) \nonumber \\
&= \frac{1}{n}\sum_{t=1}^n \mathds{E}\bigg(\frac{Y_{i,t}}{\gamma_{i,t}}\mathds{I}{\Big\{Y_{i,t} > \gamma_{i,t}B_t\Big\}}\bigg) + \frac{1}{n}\sum_{t=1}^n Z_t \label{BoundableLemma4line}
\end{align}
where $Z_t = \mathds{E}\bigg(\frac{Y_{i,t}}{\gamma_{i,t}}\mathds{I}{\Big\{Y_{i,t} \leq \gamma_{i,t}B_t\Big\}}\bigg) - \frac{Y_{i,t}}{\gamma_{i,t}}\mathds{I}{\Big\{Y_{i,t} \leq \gamma_{i,t}B_t\Big\}}$. 

We bound the first sum in (\ref{BoundableLemma4line}) by noting that \begin{displaymath}
\mathds{E}\bigg(Y_{i,t}\mathds{I}{\Big\{Y_{i,t} > \gamma_{i,t}B_t\Big\}}\bigg) \leq \mathds{E}\bigg(\frac{Y_{i,t}^2}{\gamma_{i,t}B_t}\bigg) \leq \frac{\gamma_{i,t}u_{max}}{\gamma_{i,t}B_t} = \frac{u_{max}}{B_t},
\end{displaymath}
since $\mathds{I}{\Big\{Y_{i,t} > \gamma_{i,t}B_t\Big\}} \leq \frac{Y_{i,t}}{\gamma_{i,t}B_t}$ and $\mathds{E}(Y_{i,t}^2)\leq \gamma_{i,t}u_{max}$ because $Y_{i,t} \sim Pois(\gamma_{i,t}\mu_i)$. To bound the second sum in (\ref{BoundableLemma4line}), we will use Bernstein's inequality for bounded random variables: \\

\textbf{Bernstein's Inequality:} Let $X_1,X_2,...,X_n$ be independent bounded random variables such that $\mathds{E}(X_i)=0$ and $|X_i|\leq \varsigma$ with probability 1 and let $\sigma^2 = \frac{1}{n}\sum_{i=1}^n Var(X_i)$ then for any $a>0$ we have \begin{equation}
\mathds{P}\bigg(\frac{1}{n}\sum_{i=1}^n X_i \geq a\bigg) \leq \exp \Bigg\{ \frac{-na^2}{2\sigma^2 + \frac{2\varsigma a}{3}}\Bigg\}. \label{BernsteinInequality}
\end{equation} \\

The $Z_t$ have zero mean, bounded support ($|Z_t| \leq B_t \leq B_n$), and bounded variances 

\begin{displaymath}
Var(Z_t) = Var\bigg(\frac{Y_{i,t}}{\gamma_{i,t}}\mathds{1}_{\Big\{Y_{i,t} \leq \gamma_{i,t}B_t\Big\}}\bigg) \leq \frac{1}{\gamma_{i,t}^2}\mathds{E}\bigg(Y_{i,t}^2 \mathds{1}_{\Big\{Y_{i,t} \leq \gamma_{i,t}B_t\Big\}}\bigg) \leq \frac{u_{max}}{\gamma_{i,t}}
\end{displaymath}
for $t=1,...,n$. Thus we have a bounded $\sigma^2 = \frac{1}{n}\sum_{t=1}^n Var(Z_t) \leq \frac{u_{max}}{\gamma_{min}}$ also. Therefore applying Bernstein's inequality for bounded random variables, with our upper bounds on $\sigma^2$ and $\varsigma$, have \begin{displaymath}
\mathds{P}\bigg(\frac{1}{n}\sum_{t=1}^n Z_t > a\bigg) \leq \exp \Bigg\{\frac{-na^2}{2\frac{u_{max}}{\gamma_{min}}+\frac{2B_na}{3}} \Bigg\}.
\end{displaymath}
Plugging in \begin{displaymath}
a = \sqrt{\frac{2u_{max}\ln \delta^{-1}}{\gamma_{min} n}} + \frac{B_n}{3n}\ln \delta^{-1}
\end{displaymath}
we see that $\mathds{P}\bigg(\frac{1}{n}\sum_{t=1}^n Z_t > a\bigg)  < \delta$ and therefore $\mathds{P}\bigg(\frac{1}{n}\sum_{t=1}^n Z_t \leq a\bigg)  \geq 1-\delta$. With these results we can place the following bound on (\ref{BoundableLemma4line}) that holds with at least probability $1-\delta$:
\begin{align*}
&\quad \frac{1}{n}\sum_{t=1}^n \mathds{E}\bigg(\frac{Y_{i,t}}{\gamma_{i,t}}\mathds{1}_{\Big\{Y_{i,t} > \gamma_{i,t}B_t\Big\}}\bigg) + \frac{1}{n}\sum_{t=1}^n Z_t \\
&\leq \frac{1}{n}\sum_{t=1}^n \frac{u_{max}}{\gamma_{i,t} B_t} +  \sqrt{\frac{2u_{max}\ln \delta^{-1}}{\gamma_{min} n}} + \frac{B_n}{3n}\ln \delta^{-1} \\
&= \frac{1}{n}\sum_{t=1}^n \frac{1}{\gamma_{i,t}}\sqrt{\frac{u_{max}\ln \delta^{-1}}{t}} + \sqrt{\frac{2}{\gamma_{min}}}\sqrt{\frac{u_{max}\ln \delta^{-1}}{n}} + \frac{1}{3}\sqrt{\frac{u_{max}\ln \delta^{-1}}{n}} \\
&\leq \bigg(\frac{1}{\gamma_{min}\sqrt{n}}\sum_{t=1}^n \frac{1}{\sqrt{t}} + \sqrt{\frac{2}{\gamma_{min}}} + \frac{1}{3}\bigg)\sqrt{\frac{u_{max}\ln \delta^{-1}}{n}} \\
&\leq \Big(\frac{2}{\gamma_{min}} + \sqrt{\frac{2}{\gamma_{min}}} +\frac{1}{3}\Big)\sqrt{\frac{u_{max} \ln \delta^{-1}}{n}}. \quad \quad \square
\end{align*}
This proves that \begin{displaymath}
\mathds{P}\Bigg(\mu_i \leq \hat{\mu}_{i,n}^{TruncF} + \Big(\frac{2}{\gamma_{min}} + \sqrt{\frac{2}{\gamma_{min}}} +\frac{1}{3}\Big)\sqrt{\frac{u_{max} \ln \delta^{-1}}{n}}\Bigg) \geq 1-\delta.
\end{displaymath}

\end{document}